\documentclass{article}
\usepackage{iclr2027_conference,times}

\usepackage{amsmath,amsfonts,bm}

\def\eqref#1{equation~\ref{#1}}

\def\1{\bm{1}}

\DeclareMathAlphabet{\mathsfit}{\encodingdefault}{\sfdefault}{m}{sl}
\SetMathAlphabet{\mathsfit}{bold}{\encodingdefault}{\sfdefault}{bx}{n}

\usepackage[utf8]{inputenc}
\usepackage[T1]{fontenc}
\usepackage{hyperref}
\usepackage{url}
\usepackage{booktabs}
\usepackage{tabularx}
\usepackage{amsmath}
\usepackage{amssymb}
\usepackage{amsthm}
\usepackage{graphicx}
\usepackage{xcolor}
\usepackage{xspace}
\usepackage{pifont}
\usepackage{microtype}
\usepackage{makecell}

\newcommand{\method}{answer pooling\xspace}
\newcommand{\Method}{Answer pooling\xspace}

\newtheorem{proposition}{Proposition}

\title{Your Benchmark Is Not Saturated:\\ Reviving Multiple-Choice Evaluation with \\ Answer Pooling}

\iclrfinalcopy 
\author{%
\makebox[\textwidth][c]{Mohamed Eltahir$^{1*}$\hspace{2.0em}Abobaker Ahmed$^{1*}$\hspace{2.0em}Nawaf Barebood$^{1*}$}\\[0.25em]
\makebox[\textwidth][c]{\textbf{Hussain Bu Subayt}$^{1}$\hspace{2.0em}\textbf{Tanveer Hussain}$^{2\ddagger}$\hspace{2.0em}\textbf{Naeemullah Khan}$^{1\S}$}\\[1.0em]
\makebox[\textwidth][c]{$^{1}$King Abdullah University of Science and Technology (KAUST), Thuwal, Saudi Arabia}\\
\makebox[\textwidth][c]{$^{2}$Department of Computer Science, Edge Hill University, Ormskirk, England}\\[0.3em]
\makebox[\textwidth][c]{\small\texttt{\{mohamed.hamid, nawaf.barebood, abobaker.hassan,}}\\
\makebox[\textwidth][c]{\small\texttt{hussain.busubayt, naeemullah.khan\}@kaust.edu.sa}}\\[0.1em]
\makebox[\textwidth][c]{\small\texttt{htanveer3797@gmail.com}}
}

\begin{document}

\vspace*{-30pt}
\maketitle
\lhead{Preprint.}

{\renewcommand{\thefootnote}{\fnsymbol{footnote}}
\footnotetext[1]{Equal contribution.}
\footnotetext[3]{Corresponding author.}
\footnotetext[4]{Principal Investigator (PI).}
\footnotetext{Code: https://github.com/AbobakerAhmed/AnswerPool}
}

\vspace{-20pt}

\vspace*{-0.1cm}

\begin{figure}[h]
\centering
\includegraphics[width=\linewidth]{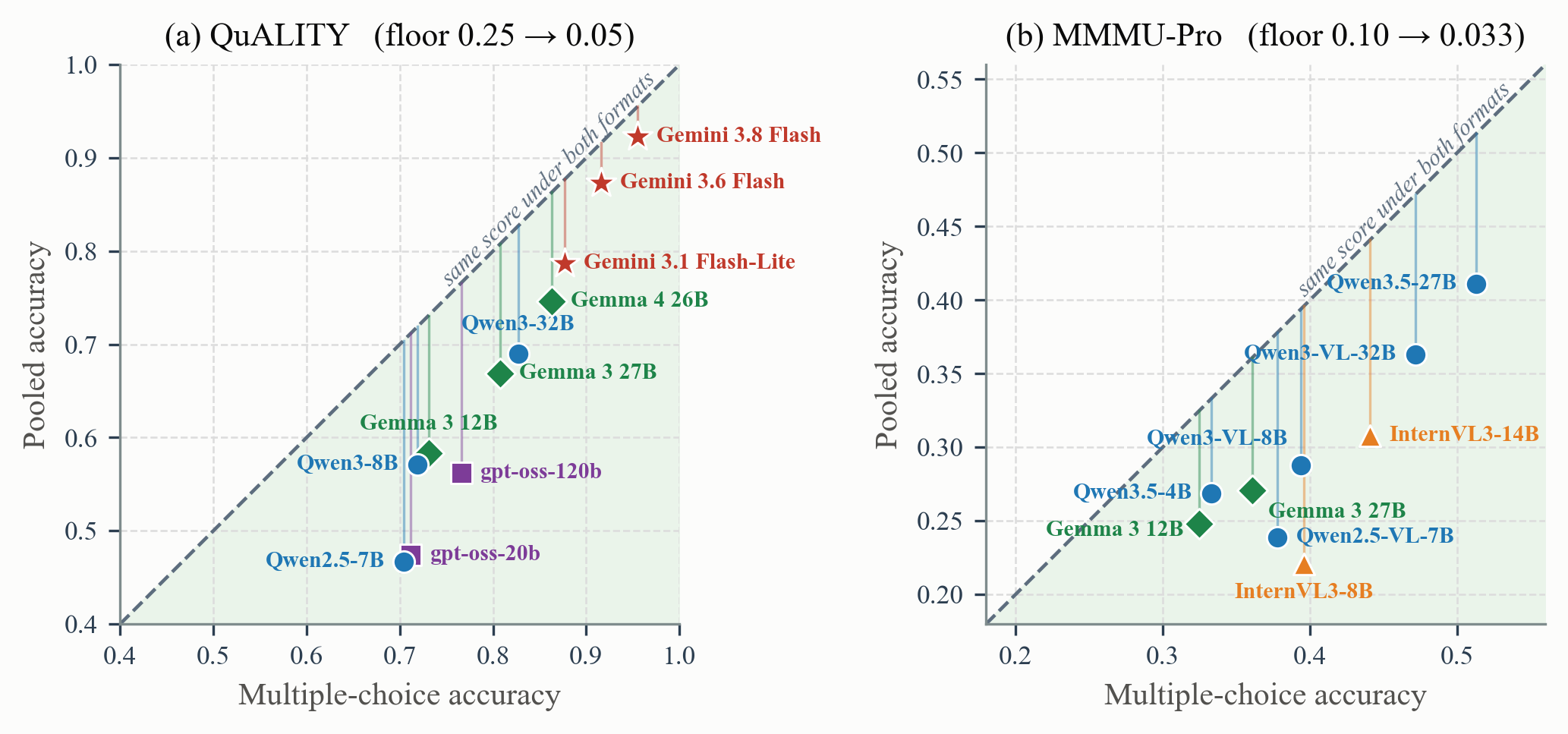}
\caption{Accuracy under multiple choice against pooled accuracy on the same
items, one point per model. Every model falls below the diagonal, and its
vertical distance to it is the elimination credit. (a) QuALITY, $N{=}5$,
$M{=}20$. (b) MMMU-Pro, $N{=}3$, $M{=}30$.}
\label{fig:scatter}
\end{figure}

\begin{abstract}
Multiple-choice benchmarks are cheap to grade and are running out of room,
and the standard remedy, writing harder items, is slow and repeated for every
benchmark. A saturated benchmark still holds a harder task. Each question's
wrong options are written for that question alone, so a model can score by
eliminating a few options. We propose \method: take $N$ questions that share
a context, pool all their options into one list, and ask the model to
assign every question its answer. No item is written and no label changes.
The chance of guessing a group right falls from $10^{-3}$ to
$5\times10^{-7}$ for five four-option questions, and a model that recognizes
its answers keeps its multiple-choice score, so the accuracy lost to pooling
measures the credit the format gave for elimination. Deleting answers from
the pool makes questions unanswerable with exact ground truth, so
abstention is scored in the same pass. Across eight text, image, and video
benchmarks and eighteen models, pooling is harder for every model, the
elimination credit is largest for the weakest models, and seven of eight
open-weight models answer 87 to 100\% of unanswerable questions. 

\end{abstract}

\section{Introduction}\label{sec:intro}

Multiple-choice question answering is the workhorse of language model
evaluation. It is cheap, needs no judge, and grades exactly, and it is
running out of room. The authors of MMLU-Pro observe that model
``performance on these benchmarks has begun to plateau, making it
increasingly difficult to discern differences in model
capabilities''~\citep{mmlupro}. The standard remedy is to write harder
items: MMLU-Pro grows four options to ten, and MMMU-Pro removes questions a
text-only model answers and adds options~\citep{mmmupro}. Each such release
is written by hand, for one benchmark, and restarts the contamination clock.


Harder items also keep the weaknesses of the format. A four-option item has
a guessing floor of 0.25. Its options carry cues of their own, and a model
shown only the choices ``bests a majority baseline in 11/12
cases''~\citep{balepur2024artifacts}. Every item also promises that its
answer is present, so measuring whether a model notices a missing answer
requires writing unanswerable items~\citep{squad2} or judging free-form
responses~\citep{abstentionbench}.

\begin{figure}[t]
\centering
\includegraphics[width=\linewidth]{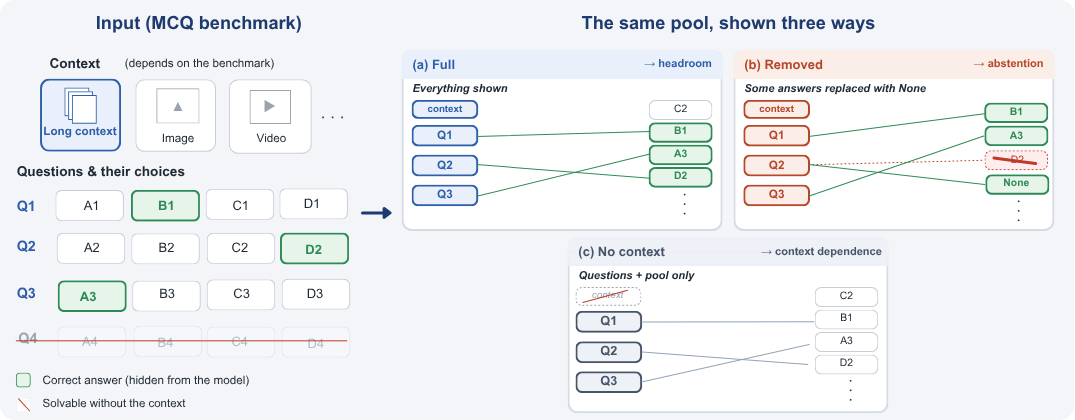}
\caption{\Method. Left: questions that share a context, each with one correct
answer (green, hidden from the model). Questions answered without the
context can be dropped first (Q4). Right: the options of the remaining
questions are pooled into one list, and the model assigns each question one
option. (a)~\textbf{Full}: every correct answer is in the pool.
(b)~\textbf{Removed answers}: some correct answers are deleted, and their
questions should be answered with none. (c)~\textbf{Removed context}: the
questions and the pool without the context.}
\label{fig:pipeline}
\end{figure}

We start from what the format discards. A question's wrong options are
written to be plausible for that question alone, so a multiple-choice item
can be answered by eliminating a few options. A benchmark that ships
several questions per passage, image, video, or topic therefore already
contains a harder task. \Method takes $N$ questions that share a context,
pools all their options into one list, and asks the model to assign every
question its answer (Figure~\ref{fig:pipeline}). Every option in the pool is
now a plausible answer to some question in the group. No item is written
and no label changes.

Two properties follow from the construction.
Each option can be used once, so guessing a whole group right has
probability $(M{-}N)!/M!$ for a pool of $M$ options, $5\times10^{-7}$ for
five four-option questions against $10^{-3}$ under multiple choice. And each
question's original item is contained in the pool, so a model that
recognizes its answer scores what it scores under multiple choice, as long
as no other option in the pool also answers the question. Accuracy
lost to pooling is therefore credit the original format gave for
eliminating a question's own options, which we call the \emph{elimination
credit}. The group size $N$ sets how much of it is withdrawn, so the same
labels can be made harder again.

The same pool gives two further read-outs without new labels. Deleting a
question's correct answer makes it unanswerable, while its own wrong options
and the answers of the other questions stay in the pool. Because the deleted
answer is known, abstention is scored exactly, without a judge, in the same
pass as accuracy. Removing the context and keeping the pool
measures how much of a score the context carries, on the same items under
both formats.

We convert eight benchmarks spanning long passages, school and expert
exams, images, and video, and evaluate eighteen models. The pooled task is harder for every model
on every benchmark (Figure~\ref{fig:scatter}). The elimination credit is
uneven: on QuALITY it is 0.03 for the strongest model and 0.24 for the
weakest, and models change rank where multiple choice paid them different
credit. Abstention separates models that accuracy does not, and seven of
the eight open-weight models answer 87 to 100\% of the questions whose
answer was removed.

\textbf{Contributions.} (1) \emph{\Method}, a label-free conversion of
grouped multiple-choice benchmarks with a combinatorial guessing floor. (2) \emph{Two read-outs from the same labels}, exact
abstention by removing answers and context dependence by removing the
context, each compared with multiple choice on identical items. (3)
\emph{Pooled builds of eight benchmarks} and an evaluation of eighteen
models showing that multiple choice pays elimination credit unevenly.

\section{Related Work}\label{sec:related}

\textbf{Benchmark saturation.} As models approach the ceiling of a
benchmark, the common response is a harder release. MMLU-Pro grows four
options to ten and adds reasoning-heavy questions~\citep{mmlupro}, MMMU-Pro
removes questions a text-only model answers and augments the
options~\citep{mmmupro}, and BeyondBench builds contamination-resistant
tests~\citep{beyondbench}. Two methods instead ask a harder question of
existing labels. CircularEval scores an item right only if the model
answers it under every rotation of its options, which lowers the floor of a
four-option item to $1/256$~\citep{mmbench}, and \citet{noneoftheothers}
replace the correct option with a none-of-the-others option to separate
reasoning from memorization. Both transform each item alone and fix its
difficulty, so neither can make the same labels harder again once models
catch up.

\textbf{Shortcuts in multiple choice.} A model can score on multiple choice
without the ability an item targets. Models beat a majority baseline from
the options alone in 11 of 12 settings~\citep{balepur2024artifacts}, prefer
options by position~\citep{zheng2024selectors}, and can often answer without
the question, which motivates free-form answers graded by a judge against a
reference~\citep{answermatching}. These studies diagnose the format or leave
it. \Method keeps exact grading and changes what the options offer: each
question's options are mixed with the plausible answers of its neighbours.

\textbf{Unanswerable questions.} Measuring whether a model declines requires
items without an answer. SQuAD~2.0 writes them by hand~\citep{squad2},
AbstentionBench curates twenty datasets and finds that scale does not fix
abstention~\citep{abstentionbench}, and none-of-the-above studies add a
printed escape option to every item~\citep{madhusudhan2024abstain,
tam2025nota}. Each relies on authored labels, a judge, or a single format.
Deleting an answer from a pool creates the unanswerable item from labels the
benchmark already has and scores it next to accuracy.

Together these threads leave one question open: whether the labels a
benchmark already ships can be asked a harder question, one that can be made
harder again. Section~\ref{sec:method}
builds that question from the groups benchmarks already contain.

\section{\Method}\label{sec:method}

\Method turns the items of a benchmark into groups and asks one assignment
question per group. We define the pool and its scores
(Section~\ref{sec:pool}), state what the pool guarantees
(Section~\ref{sec:properties}), derive two read-outs by removing answers or
the context (Section~\ref{sec:modes}), and describe how a pooled group is
kept valid (Section~\ref{sec:validity}).

\subsection{Groups and Pools}\label{sec:pool}

A \textbf{group} is a context $c$ and $N$ questions $q_1,\dots,q_N$ about it.
The context is a passage, an image, or a video. For benchmarks without a
context, a group is $N$ questions on the same topic and $c$ is empty.
Question $q_i$ has one correct answer $g_i$ and a set $D_i$ of $k{-}1$ wrong
options, so its multiple-choice item is $(q_i,\{g_i\}\cup D_i)$, with
guessing floor $1/k$.

The \textbf{pool} of a group is the set of all its options in random order,
\begin{equation}
P=\mathrm{shuffle}\Big(\bigcup_{i=1}^{N}\big(\{g_i\}\cup D_i\big)\Big),
\qquad M=|P|,
\label{eq:pool}
\end{equation}
with identical options merged, so $M=kN$ when no two options coincide. The
model reads $c$, the $N$ questions, and the options of $P$, each with a
letter label, and returns an \textbf{assignment} $a:\{1,\dots,N\}\to P$ that
uses each option at most once. The correct assignment is $a^\star(i)=g_i$.

Two scores compare the pooled task with multiple choice on the same items.
\textbf{Accuracy} is the fraction of questions assigned their correct
answer, $|\{i : a(i)=g_i\}|/N$, and is compared with multiple-choice
accuracy on the same questions. The
\textbf{all-correct rate} is the fraction of groups whose assignment is
fully correct, $a=a^\star$, and is compared with the rate at which all $N$
multiple-choice items of a group are answered correctly.

\subsection{Guessing Floor and Recoverability}\label{sec:properties}

\begin{proposition}[Guessing floor]\label{prop:floor}
An assignment drawn uniformly from those that use each option at most once
is correct with probability $(M-N)!/M!$.
\end{proposition}

For $N$ independent $k$-option items the all-correct floor is $k^{-N}$. At
$N{=}5$ and $k{=}4$ it is $9.8\times10^{-4}$, and the pool of $M{=}20$ options
lowers it to $5.4\times10^{-7}$. The per-question floor falls from $1/k$ to
$1/M$.

\begin{proposition}[Multiple choice is recoverable]\label{prop:recover}
Suppose that no option in $P$ other than $g_i$ answers $q_i$. A model that
selects the option answering $q_i$ selects $g_i$ under both formats, so its
accuracy on $q_i$ is the same under multiple choice and pooling.
\end{proposition}

Pooling adds candidates and removes none. A model that recognizes its
answers loses nothing. A model that picks $g_i$ because it is the least
implausible of the $k$ options written for $q_i$ can lose it to the answers
of the other questions. We call the difference between multiple-choice
accuracy and pooled accuracy on the same questions the \textbf{elimination
credit}. The group size $N$ controls how much of it is withdrawn: a larger
group puts more plausible answers in the pool and lowers the floor of
Proposition~\ref{prop:floor}, so a benchmark can be made harder again
without new labels.

\subsection{Removing Answers and Removing the Context}\label{sec:modes}

The same pool yields two further read-outs, and neither needs new labels.

\textbf{Removed answers.} We choose a set $W\subset\{1,\dots,N\}$ of
$\mathrm{round}(pN)$ questions for a fraction $p$, delete their correct
answers from $P$, and set $a^\star(i)=\mathrm{none}$ for $i\in W$. The wrong
options of these questions and the answers of the others stay in $P$. The
assignment may now map a question to none, and the prompt permits it without
saying how often. Because $W$ is known, the ground truth is exact and needs
no judge. The \textbf{false-answer rate} is the fraction of questions in $W$
that the model answers anyway, and the \textbf{false-abstention rate} is the
fraction of the remaining questions that it declines. Accuracy is computed
on the remaining questions. We compare both rates with
two multiple-choice versions of the same removed set. \textbf{MCQ with a none
option} prints ``none of these'' as one option of every item, and
\textbf{MCQ with an instruction} deletes the answer from the item and
permits none in the instruction. Both hold the number of options fixed, so
the count does not reveal whether the answer was removed.

\textbf{Removed context.} The model sees the questions and the pool without
$c$, and the same questions are asked as multiple-choice items without $c$.
The share of accuracy each format loses measures how much of its score the
context carries.

\subsection{Validity}\label{sec:validity}

The source benchmark validated each question against its own options.
Proposition~\ref{prop:recover} needs one further condition, that no other
option in the pool also answers $q_i$. We screen for it: a model sees the
context, one question, the pool, and the question's correct answer, and
names any other option that also answers the question. A question is flagged
when two of three screening models name one, and Section~\ref{sec:ablations}
compares results on all groups and on the groups with no flagged question.
Where a benchmark has enough questions per context, we also remove before
conversion every question that two of three models answer correctly without
the context, so that the retained questions need it.

\section{Experiments}\label{sec:experiments}

\subsection{Setup}\label{sec:setup}

\textbf{Benchmarks.} We convert four text benchmarks and three image and
video benchmarks, and report C-Eval~\citep{ceval} in
Appendix~\ref{app:permodel}. QuALITY~\citep{quality} and RACE~\citep{race}
group questions by passage, MMLU-Pro~\citep{mmlupro} and GPQA~\citep{gpqa}
by topic, MMMU-Pro~\citep{mmmupro} by subject, MedXpertQA-MM~\citep{medxpertqa}
by body system, and Video-MME~\citep{videomme} by video. Every wrong option
of every question enters the pool, and image questions keep their images.
QuALITY is first filtered to questions that need the passage. Filtering RACE
would leave too few questions per passage, so we keep it whole and measure
its passage dependence in Section~\ref{sec:context}. Construction details
are in
Appendix~\ref{app:construction}.

\begin{table}[t]
\centering
\small
\caption{Text benchmarks. Accuracy under multiple choice / pooled on
identical items. Footer: group and pool sizes, the all-correct rate of the
best model, the all-correct guessing floor, and the rank correlation of
models between the two formats.}
\label{tab:text}
\setlength{\tabcolsep}{4pt}
\vspace{5pt}
\begin{tabular}{lcccc}
\toprule
\textbf{Model} & QuALITY & RACE & MMLU-Pro & GPQA \\
\midrule
Gemini 3.8 Flash      & 0.955 / 0.923 & 0.963 / 0.929 & 0.898 / 0.886 & 0.887 / 0.812 \\
Gemini 3.6 Flash      & 0.916 / 0.873 & 0.958 / 0.863 & 0.776 / 0.722 & 0.648 / 0.566 \\
Gemini 3.1 Flash-Lite & 0.877 / 0.787 & 0.937 / 0.821 & 0.691 / 0.597 & 0.542 / 0.453 \\
Gemma 4 26B           & 0.863 / 0.746 & 0.918 / 0.768 & 0.725 / 0.518 & 0.511 / 0.376 \\
Gemma 3 27B           & 0.808 / 0.669 & 0.897 / 0.745 & 0.478 / 0.350 & 0.357 / 0.308 \\
Gemma 3 12B           & 0.731 / 0.583 & 0.882 / 0.716 & 0.424 / 0.312 & 0.311 / 0.299 \\
Qwen3-32B             & 0.827 / 0.690 & 0.916 / 0.795 & 0.561 / 0.442 & 0.424 / 0.354 \\
Qwen3-8B              & 0.719 / 0.571 & 0.866 / 0.689 & 0.427 / 0.314 & 0.388 / 0.272 \\
Qwen2.5-7B            & 0.704 / 0.467 & 0.895 / 0.637 & 0.426 / 0.250 & 0.316 / 0.234 \\
gpt-oss-120b          & 0.766 / 0.562 & 0.882 / 0.755 & 0.765 / 0.687 & 0.655 / 0.508 \\
gpt-oss-20b           & 0.712 / 0.474 & 0.879 / 0.676 & 0.681 / 0.552 & 0.557 / 0.407 \\
\midrule
$N$ / $M$              & 5 / 20 & 4 / 16 & 3 / 30 & 5 / 20 \\
All-correct, best model & 0.787 / \textbf{0.683} & 0.863 / \textbf{0.747} & 0.743 / \textbf{0.714} & 0.554 / \textbf{0.349} \\
All-correct floor      & $10^{-3}$ / $5{\times}10^{-7}$ & $4{\times}10^{-3}$ / $2{\times}10^{-5}$ & $10^{-3}$ / $4{\times}10^{-5}$ & $10^{-3}$ / $5{\times}10^{-7}$ \\
Rank correlation $\rho$ & 0.97 & 0.87 & 0.96 & 0.94 \\
\bottomrule
\end{tabular}
\end{table}

\textbf{Models.} Eleven text models: Gemini 3.8 Flash, Gemini 3.6 Flash,
Gemini 3.1 Flash-Lite, and Gemma 4 26B through Google's API, and Gemma 3 12B
and 27B, Qwen2.5-7B, Qwen3-8B, Qwen3-32B, gpt-oss-20b, and gpt-oss-120b
served locally. Nine vision models: Gemma 3 12B and 27B, Qwen2.5-VL-7B,
Qwen3-VL-8B and 32B, InternVL3-8B and 14B, and Qwen3.5-4B and 27B.

\textbf{Protocol.} All models answer directly, with thinking disabled where
the chat template allows it and the lowest reasoning effort for gpt-oss, so
absolute scores sit below the thinking-mode scores of model cards. Decoding
is greedy, and local models decode under a grammar that admits only
well-formed answers. Multiple choice and pooling are always compared on
identical items. A second
build seed moves accuracy and the all-correct rate by at most 0.03, so we do not interpret smaller differences.
Protocol details and prompts are in
Appendices~\ref{app:protocol} and~\ref{app:prompts}.

\subsection{The Same Labels Become a Harder Task}\label{sec:main}

Every model
loses accuracy on every benchmark when its options are pooled
(Tables~\ref{tab:text} and~\ref{tab:vision}), and the all-correct rate falls
further. On QuALITY the strongest model answers 79\% of groups fully
correctly under multiple choice and 68\% when pooled, against a guessing
floor a thousand times lower. MMMU-Pro lowered its floor to 0.10 by writing
ten options per item~\citep{mmmupro}. Pooling the same options lowers the
all-correct floor to $4\times10^{-5}$ and reduces the accuracy of every
vision model by 0.06 to 0.18.

\begin{figure}[!h]
\centering
\includegraphics[width=0.85\linewidth]{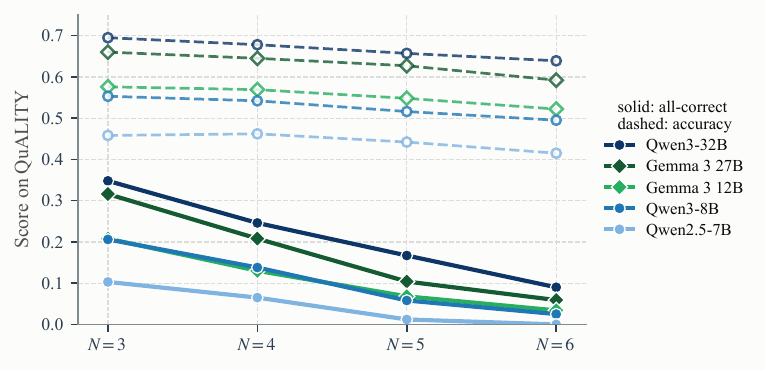}
\caption{Group size $N$ on QuALITY for five models. All-correct
rate (solid) and accuracy (dashed).}
\label{fig:nknob}
\end{figure}

Raising $N$ from 3 to 6
on QuALITY lowers every model's all-correct rate while accuracy changes
little (Figure~\ref{fig:nknob}). Qwen3-32B falls from 0.35 to 0.09 on whole
groups and from 0.70 to 0.64 per question, and the five models keep their
order at every $N$. The same labels can therefore be
released at increasing difficulty as models improve.

\subsection{The Drop Is the Elimination Credit}\label{sec:credit}

\begin{table}[t]
\centering
\small
\caption{Image and video benchmarks, same layout as Table~\ref{tab:text}.}
\label{tab:vision}
\setlength{\tabcolsep}{4pt}
\vspace{5pt}
\begin{tabular}{lccc}
\toprule
\textbf{Model} & MedXpertQA-MM & MMMU-Pro & Video-MME \\
\midrule
Qwen3.5-27B    & 0.429 / 0.349 & 0.513 / 0.411 & 0.690 / 0.628 \\
Qwen3.5-4B     & 0.282 / 0.225 & 0.333 / 0.269 & 0.561 / 0.504 \\
Qwen3-VL-32B   & 0.298 / 0.259 & 0.472 / 0.363 & 0.647 / 0.595 \\
Qwen3-VL-8B    & 0.252 / 0.205 & 0.394 / 0.288 & 0.609 / 0.537 \\
Qwen2.5-VL-7B  & 0.227 / 0.188 & 0.378 / 0.239 & 0.577 / 0.473 \\
Gemma 3 27B    & 0.240 / 0.215 & 0.361 / 0.271 & 0.627 / 0.548 \\
Gemma 3 12B    & 0.234 / 0.203 & 0.325 / 0.248 & 0.592 / 0.499 \\
InternVL3-14B  & 0.255 / 0.191 & 0.441 / 0.308 & 0.650 / 0.563 \\
InternVL3-8B   & 0.241 / 0.184 & 0.396 / 0.221 & 0.620 / 0.483 \\
\midrule
$N$ / $M$              & 3 / 15 & 3 / 30 & 3 / 12 \\
All-correct, best model & 0.063 / \textbf{0.044} & 0.162 / \textbf{0.090} & 0.356 / \textbf{0.281} \\
All-correct floor      & $8{\times}10^{-3}$ / $4{\times}10^{-4}$ & $10^{-3}$ / $4{\times}10^{-5}$ & $2{\times}10^{-2}$ / $8{\times}10^{-4}$ \\
Rank correlation $\rho$ & 0.73 & 0.65 & 0.82 \\
\bottomrule
\end{tabular}
\end{table}

On QuALITY the credit
ranges from 0.03 for Gemini 3.8 Flash to 0.24 for Qwen2.5-7B and
gpt-oss-20b. The six models above 0.80 under multiple choice lose at most
0.14, the five below 0.77 lose 0.15 to 0.24, and the accuracy gap between
the best and the worst model grows from 0.25 to 0.46. On MMLU-Pro the
strongest model loses 0.01. Strong models mostly recognize their answers,
and weaker models were credited for elimination.

The rank correlation
between the two formats is 0.87 to 0.97 on text and 0.65 to 0.82 on images
and video. Models change places when they are close under multiple choice
and carry different credit. On QuALITY, gpt-oss-120b leads Gemma 3 12B by
0.035 under multiple choice, carries a credit of 0.20 against 0.15, and
falls below it. On MMMU-Pro, InternVL3-8B carries a credit of 0.18 and falls
from fourth to last.

A pooled prompt is longer and asks
for a full assignment, so the drop could be a cost of the format. An
unrelated pool tests this. It keeps each group's questions, correct answers,
prompt, and pool size, and replaces every wrong option with a correct answer
of another group. Every text model then scores at or above its
multiple-choice accuracy on QuALITY, and so do seven vision models on
MMMU-Pro (Table~\ref{tab:unrelated}). A model
therefore loses accuracy only when plausible wrong options share the pool,
and related questions supply part of them: in cross-domain groups, which mix
questions from different passages, every model on RACE recovers part of its
loss, a third on average (Table~\ref{tab:crossdomain}). Pooled accuracy
thus measures how well a model recognizes its answers among plausible
alternatives, and grouping questions that share a context supplies more of
them.

\subsection{Abstention}\label{sec:abstention}

\begin{table}[t]
\centering
\small
\caption{Abstention on QuALITY with two of five answers removed per group,
identical questions and removed answers in every column. Acc.: pooled
accuracy on the answerable questions.}
\label{tab:abstention}
\setlength{\tabcolsep}{5pt}
\vspace{5pt}
\begin{tabular}{lc cc cc}
\toprule
 & & \multicolumn{2}{c}{False answer $\downarrow$} & \multicolumn{2}{c}{False abstention $\downarrow$} \\
\textbf{Model} & Acc. & Pooled & MCQ+none & Pooled & MCQ+none \\
\midrule
Gemini 3.8 Flash      & 0.758 & \textbf{0.167} & \textbf{0.103} & 0.203 & 0.180 \\
Gemini 3.6 Flash      & 0.750 & 0.278 & 0.120 & 0.144 & 0.270 \\
Gemini 3.1 Flash-Lite & 0.753 & 0.698 & 0.242 & 0.046 & 0.164 \\
Gemma 4 26B           & 0.690 & 0.677 & 0.337 & 0.084 & 0.139 \\
Gemma 3 27B           & 0.626 & 0.922 & 0.553 & 0.021 & 0.087 \\
Gemma 3 12B           & 0.536 & 0.970 & 0.563 & 0.016 & 0.146 \\
Qwen3-32B             & 0.684 & 0.993 & 0.522 & 0.000 & 0.109 \\
Qwen3-8B              & 0.519 & 0.998 & 0.457 & 0.000 & 0.189 \\
Qwen2.5-7B            & 0.427 & 0.945 & 0.380 & 0.032 & 0.268 \\
gpt-oss-120b          & 0.547 & 0.877 & 0.440 & 0.026 & 0.131 \\
gpt-oss-20b           & 0.464 & 0.873 & 0.515 & 0.059 & 0.136 \\
\bottomrule
\end{tabular}
\end{table}

Table~\ref{tab:abstention}
removes two of five answers in every QuALITY group. The false-answer rate
spans 0.17 to 1.00 across models whose accuracy spans 0.43 to 0.76, and the
two do not share an order: Gemini 3.6 Flash and Gemini 3.1 Flash-Lite
differ by 0.003 in accuracy and by 0.42 in false-answer rate. Seven of the
eight open-weight models answer 87 to 100\% of the unanswerable questions at
every size from 7B to 120B, and on MedXpertQA-MM the nine vision models
answer 96 to 100\%. The false-answer rate of the strongest model rises with
the domain, from 0.17 on QuALITY to 0.38 on GPQA and 0.54 on MedXpertQA-MM
(Appendix~\ref{app:abstention}).

On the same questions and
removed answers, a printed none option lowers the false-answer rate of every
model, to 0.10 to 0.56, and raises the false-abstention rate of ten of the
eleven. The pooled rate measures abstention without a printed escape, and
any abstention rate has to be reported with the format that produced it.

\subsection{Context Dependence}\label{sec:context}

\begin{table}[t]
\centering
\small
\caption{Share of accuracy lost when the context is removed, mean over
models, on identical items under both formats. QuALITY: the eight models not
used by its no-context filter.}
\label{tab:context}
\setlength{\tabcolsep}{6pt}
\vspace{5pt}
\begin{tabular}{llcc}
\toprule
\textbf{Benchmark} & \textbf{Context} & MCQ & Pooled \\
\midrule
QuALITY   & passage & 56\% & 49\% \\
RACE      & passage & 43\% & 37\% \\
MMMU-Pro  & images  & 35\% & 36\% \\
Video-MME & video   & 34\% & 27\% \\
\bottomrule
\end{tabular}
\end{table}

Removing the context
costs both formats a large share of their accuracy on the same items
(Table~\ref{tab:context}). RACE is known to be partly answerable without its
passages~\citep{si2019bert}, and it loses 43\% of its accuracy under
multiple choice and 37\% when pooled. How much a score depends on the
context is therefore a property of the benchmark that its pooled build keeps.
QuALITY, filtered to questions that need the passage, loses about half under
both formats.

\subsection{Ablations}\label{sec:ablations}


\begin{table}[t]
\centering
\caption{Protocol changes on QuALITY. Largest change over the five local
models, in the score that moves most, Qwen3-8B for the grammar.}
\vspace{5pt}
\label{tab:stability}
\small
\setlength{\tabcolsep}{6pt}
\begin{tabular}{llc}
\toprule
Change & Score & Largest change \\
\midrule
Second build seed                         & all-correct  & 0.027 \\
Second wording of the decline instruction & false answer & 0.028 \\
Options may be used twice                 & false answer & 0.014 \\
No output grammar                         & false answer & 0.004 \\
\bottomrule
\end{tabular}
\end{table}

\begin{table}[t]
\begin{minipage}[t]{0.46\linewidth}
\centering
\caption{Unrelated pool: every wrong option is replaced by a correct answer
of another group. Mean accuracy of the five local models on QuALITY and
seven vision models on MMMU-Pro.}
\vspace{5pt}
\label{tab:unrelated}
\small
\setlength{\tabcolsep}{4pt}
\begin{tabular}{lcc}
\toprule
Format & QuALITY & MMMU-Pro \\
\midrule
Multiple choice & 0.758 & 0.395 \\
Pooled          & 0.596 & 0.277 \\
Unrelated pool  & \textbf{0.842} & \textbf{0.798} \\
\bottomrule
\end{tabular}
\end{minipage}\hfill
\begin{minipage}[t]{0.51\linewidth}
\centering
\caption{Cross-domain groups mix questions from different passages or
subdomains, each with its own options. Mean accuracy of the five local
models.}
\vspace{5pt}
\label{tab:crossdomain}
\small
\setlength{\tabcolsep}{4pt}
\begin{tabular}{lccc}
\toprule
Format & RACE & GPQA & QuALITY \\
\midrule
Multiple choice      & 0.891 & 0.359 & 0.758 \\
Pooled, same context & 0.716 & 0.293 & 0.596 \\
Pooled, cross-domain & 0.775 & 0.317 & 0.387 \\
\bottomrule
\end{tabular}
\end{minipage}
\end{table}

A second build seed, a
second wording of the decline instruction, reusable options, and decoding
without the output grammar each move every score by less than 0.03
(Table~\ref{tab:stability}). On the groups the ambiguity check keeps, the
all-correct rate of Gemini 3.6 Flash rises from 0.520 to 0.562, and its
false-answer rate moves by 0.08 as the removed fraction grows from 0.2 to
0.8 (Appendix~\ref{app:controls}).

Every wrong option must enter the
pool: with 3 of the 9 wrong options of each MMMU-Pro question, the pooled
task is easier than the ten-option item, and it becomes harder as the rest
are added (Appendix~\ref{app:controls}). On long passages, groups must share
a context: a cross-domain group on QuALITY needs five articles in one
prompt, and accuracy falls from 0.60 to 0.39 (Table~\ref{tab:crossdomain}).
\section{Conclusion and Limitations}\label{sec:conclusion}

A saturated multiple-choice benchmark still holds a harder task in the labels
it ships. \Method pools the options of questions that share a context,
withdraws the credit that multiple choice gives for elimination, and lowers
the chance of guessing a group right by orders of magnitude, with no new
annotation. The same pool scores abstention exactly, and a larger group
makes the task harder again as models improve. 

\textbf{Limitations.} The conversion needs benchmarks that group several
questions by context or topic. Proposition~\ref{prop:recover} assumes that
no pooled option answers two questions, which we check with screening
models. A pooled prompt also holds every question of a group with
all their options, and on image benchmarks all their images, so it is longer
than a single item. The unrelated pool shows that this costs no accuracy at
the sizes we use, and a group shares one context so that no prompt holds
several passages, but larger groups lengthen the prompt further which might add long-context as an indirect factor to be considered.



\subsubsection*{Acknowledgments}
We are grateful to the KAUST Academy for its generous support, and especially to Prof. Sultan
Albarakati who made this work possible. For computer time, this research used Ibex managed by the
Supercomputing Core Laboratory at King Abdullah University of Science \& Technology (KAUST)
in Thuwal, Saudi Arabia.

\bibliography{bibliography}

\begin{thebibliography}{19}
\providecommand{\natexlab}[1]{#1}
\providecommand{\url}[1]{\texttt{#1}}
\expandafter\ifx\csname urlstyle\endcsname\relax
  \providecommand{\doi}[1]{doi: #1}\else
  \providecommand{\doi}{doi: \begingroup \urlstyle{rm}\Url}\fi

\bibitem[Balepur et~al.(2024)Balepur, Ravichander, and Rudinger]{balepur2024artifacts}
Nishant Balepur, Abhilasha Ravichander, and Rachel Rudinger.
\newblock Artifacts or abduction: How do {LLM}s answer multiple-choice questions without the question?
\newblock In Lun-Wei Ku, Andre Martins, and Vivek Srikumar (eds.), \emph{Proceedings of the 62nd Annual Meeting of the Association for Computational Linguistics (Volume 1: Long Papers)}, pp.\  10308--10330, Bangkok, Thailand, August 2024. Association for Computational Linguistics.
\newblock \doi{10.18653/v1/2024.acl-long.555}.
\newblock URL \url{https://aclanthology.org/2024.acl-long.555/}.

\bibitem[Chandak et~al.(2025)Chandak, Goel, Prabhu, Hardt, and Geiping]{answermatching}
Nikhil Chandak, Shashwat Goel, Ameya Prabhu, Moritz Hardt, and Jonas Geiping.
\newblock Answer matching outperforms multiple choice for language model evaluation.
\newblock \emph{arXiv preprint arXiv:2507.02856}, 2025.

\bibitem[Fu et~al.(2025)Fu, Dai, Luo, Li, Ren, Zhang, Wang, Zhou, Shen, Zhang, et~al.]{videomme}
Chaoyou Fu, Yuhan Dai, Yongdong Luo, Lei Li, Shuhuai Ren, Renrui Zhang, Zihan Wang, Chenyu Zhou, Yunhang Shen, Mengdan Zhang, et~al.
\newblock Video-mme: The first-ever comprehensive evaluation benchmark of multi-modal llms in video analysis.
\newblock In \emph{2025 IEEE/CVF Conference on Computer Vision and Pattern Recognition (CVPR)}, pp.\  24108--24118. IEEE, 2025.

\bibitem[Huang et~al.(2023)Huang, Bai, Zhu, Zhang, Zhang, Su, Liu, Lv, Zhang, Fu, et~al.]{ceval}
Yuzhen Huang, Yuzhuo Bai, Zhihao Zhu, Junlei Zhang, Jinghan Zhang, Tangjun Su, Junteng Liu, Chuancheng Lv, Yikai Zhang, Yao Fu, et~al.
\newblock C-eval: A multi-level multi-discipline chinese evaluation suite for foundation models.
\newblock \emph{Advances in neural information processing systems}, 36:\penalty0 62991--63010, 2023.

\bibitem[Kirichenko et~al.(2026)Kirichenko, Ibrahim, Chaudhuri, and Bell]{abstentionbench}
Polina Kirichenko, Mark Ibrahim, Kamalika Chaudhuri, and Samuel~J Bell.
\newblock Abstentionbench: Reasoning llms fail on unanswerable questions.
\newblock \emph{Advances in Neural Information Processing Systems}, 38, 2026.

\bibitem[Lai et~al.(2017)Lai, Xie, Liu, Yang, and Hovy]{race}
Guokun Lai, Qizhe Xie, Hanxiao Liu, Yiming Yang, and Eduard Hovy.
\newblock {RACE}: Large-scale {R}e{A}ding comprehension dataset from examinations.
\newblock In Martha Palmer, Rebecca Hwa, and Sebastian Riedel (eds.), \emph{Proceedings of the 2017 Conference on Empirical Methods in Natural Language Processing}, pp.\  785--794, Copenhagen, Denmark, September 2017. Association for Computational Linguistics.
\newblock \doi{10.18653/v1/D17-1082}.
\newblock URL \url{https://aclanthology.org/D17-1082/}.

\bibitem[Liu et~al.(2024)Liu, Duan, Zhang, Li, Zhang, Zhao, Yuan, Wang, He, Liu, et~al.]{mmbench}
Yuan Liu, Haodong Duan, Yuanhan Zhang, Bo~Li, Songyang Zhang, Wangbo Zhao, Yike Yuan, Jiaqi Wang, Conghui He, Ziwei Liu, et~al.
\newblock Mmbench: Is your multi-modal model an all-around player?
\newblock In \emph{European conference on computer vision}, pp.\  216--233. Springer, 2024.

\bibitem[Madhusudhan et~al.(2025)Madhusudhan, Madhusudhan, Yadav, and Hashemi]{madhusudhan2024abstain}
Nishanth Madhusudhan, Sathwik~Tejaswi Madhusudhan, Vikas Yadav, and Masoud Hashemi.
\newblock Do llms know when to not answer? investigating abstention abilities of large language models.
\newblock In \emph{Proceedings of the 31st International Conference on Computational Linguistics}, pp.\  9329--9345, 2025.

\bibitem[Pang et~al.(2022)Pang, Parrish, Joshi, Nangia, Phang, Chen, Padmakumar, Ma, Thompson, He, and Bowman]{quality}
Richard~Yuanzhe Pang, Alicia Parrish, Nitish Joshi, Nikita Nangia, Jason Phang, Angelica Chen, Vishakh Padmakumar, Johnny Ma, Jana Thompson, He~He, and Samuel~R. Bowman.
\newblock {Q}u{ALITY}: Question answering with long input texts, yes!
\newblock In Marine Carpuat, Marie-Catherine de~Marneffe, and Ivan~Vladimir Meza~Ruiz (eds.), \emph{Proceedings of the 2022 Conference of the North American Chapter of the Association for Computational Linguistics: Human Language Technologies}, pp.\  5336--5358, Seattle, United States, July 2022. Association for Computational Linguistics.
\newblock \doi{10.18653/v1/2022.naacl-main.391}.
\newblock URL \url{https://aclanthology.org/2022.naacl-main.391/}.

\bibitem[Rajpurkar et~al.(2018)Rajpurkar, Jia, and Liang]{squad2}
Pranav Rajpurkar, Robin Jia, and Percy Liang.
\newblock Know what you don{'}t know: Unanswerable questions for {SQ}u{AD}.
\newblock In Iryna Gurevych and Yusuke Miyao (eds.), \emph{Proceedings of the 56th Annual Meeting of the Association for Computational Linguistics (Volume 2: Short Papers)}, pp.\  784--789, Melbourne, Australia, July 2018. Association for Computational Linguistics.
\newblock \doi{10.18653/v1/P18-2124}.
\newblock URL \url{https://aclanthology.org/P18-2124/}.

\bibitem[Rein et~al.(2023)Rein, Hou, Stickland, Petty, Pang, Dirani, Michael, and Bowman]{gpqa}
David Rein, Betty~Li Hou, Asa~Cooper Stickland, Jackson Petty, Richard~Yuanzhe Pang, Julien Dirani, Julian Michael, and Samuel~R Bowman.
\newblock Gpqa: A graduate-level google-proof q\&a benchmark.
\newblock \emph{arXiv preprint arXiv:2311.12022}, 2023.

\bibitem[Salido et~al.(2025)Salido, Gonzalo, and Marco]{noneoftheothers}
Eva~S{\'a}nchez Salido, Julio Gonzalo, and Guillermo Marco.
\newblock None of the others: a general technique to distinguish reasoning from memorization in multiple-choice llm evaluation benchmarks.
\newblock \emph{arXiv preprint arXiv:2502.12896}, 2025.

\bibitem[Si et~al.(2019)Si, Wang, Kan, and Jiang]{si2019bert}
Chenglei Si, Shuohang Wang, Min-Yen Kan, and Jing Jiang.
\newblock What does bert learn from multiple-choice reading comprehension datasets?
\newblock \emph{arXiv preprint arXiv:1910.12391}, 2019.

\bibitem[Srivastava et~al.(2026)Srivastava, Hussain, Bi, Roy, Pitre, Lu, Ziyadi, and Wang]{beyondbench}
Gaurav Srivastava, Aafiya Hussain, Zhenyu Bi, Swastik Roy, Priya Pitre, Meng Lu, Morteza Ziyadi, and Xuan Wang.
\newblock Beyondbench: Contamination-resistant evaluation of reasoning in language models.
\newblock In \emph{International Conference on Learning Representations}, volume 2026, pp.\  20501--20636, 2026.

\bibitem[Tam et~al.(2025)Tam, Wu, Lin, and Chen]{tam2025nota}
Zhi~Rui Tam, Cheng-Kuang Wu, Chieh-Yen Lin, and Yun-Nung Chen.
\newblock None of the above, less of the right parallel patterns in human and {LLM} performance on multi-choice questions answering.
\newblock In Wanxiang Che, Joyce Nabende, Ekaterina Shutova, and Mohammad~Taher Pilehvar (eds.), \emph{Findings of the Association for Computational Linguistics: ACL 2025}, pp.\  20112--20134, Vienna, Austria, July 2025. Association for Computational Linguistics.
\newblock ISBN 979-8-89176-256-5.
\newblock \doi{10.18653/v1/2025.findings-acl.1031}.
\newblock URL \url{https://aclanthology.org/2025.findings-acl.1031/}.

\bibitem[Wang et~al.(2024)Wang, Ma, Zhang, Ni, Chandra, Guo, Ren, Arulraj, He, Jiang, et~al.]{mmlupro}
Yubo Wang, Xueguang Ma, Ge~Zhang, Yuansheng Ni, Abhranil Chandra, Shiguang Guo, Weiming Ren, Aaran Arulraj, Xuan He, Ziyan Jiang, et~al.
\newblock Mmlu-pro: A more robust and challenging multi-task language understanding benchmark.
\newblock \emph{Advances in Neural Information Processing Systems}, 37:\penalty0 95266--95290, 2024.

\bibitem[Yue et~al.(2025)Yue, Zheng, Ni, Wang, Zhang, Tong, Sun, Yu, Zhang, Sun, Su, Chen, and Neubig]{mmmupro}
Xiang Yue, Tianyu Zheng, Yuansheng Ni, Yubo Wang, Kai Zhang, Shengbang Tong, Yuxuan Sun, Botao Yu, Ge~Zhang, Huan Sun, Yu~Su, Wenhu Chen, and Graham Neubig.
\newblock {MMMU}-pro: A more robust multi-discipline multimodal understanding benchmark.
\newblock In Wanxiang Che, Joyce Nabende, Ekaterina Shutova, and Mohammad~Taher Pilehvar (eds.), \emph{Proceedings of the 63rd Annual Meeting of the Association for Computational Linguistics (Volume 1: Long Papers)}, pp.\  15134--15186, Vienna, Austria, July 2025. Association for Computational Linguistics.
\newblock ISBN 979-8-89176-251-0.
\newblock \doi{10.18653/v1/2025.acl-long.736}.
\newblock URL \url{https://aclanthology.org/2025.acl-long.736/}.

\bibitem[Zheng et~al.(2024)Zheng, Zhou, Meng, Zhou, and Huang]{zheng2024selectors}
Chujie Zheng, Hao Zhou, Fandong Meng, Jie Zhou, and Minlie Huang.
\newblock Large language models are not robust multiple choice selectors.
\newblock In \emph{International Conference on Learning Representations}, volume 2024, pp.\  19426--19454, 2024.

\bibitem[Zuo et~al.(2025)Zuo, Qu, Li, Chen, Zhu, Hua, Zhang, Ding, and Zhou]{medxpertqa}
Yuxin Zuo, Shang Qu, Yifei Li, Zhangren Chen, Xuekai Zhu, Ermo Hua, Kaiyan Zhang, Ning Ding, and Bowen Zhou.
\newblock Medxpertqa: Benchmarking expert-level medical reasoning and understanding.
\newblock \emph{arXiv preprint arXiv:2501.18362}, 2025.

\end{thebibliography}
\bibliographystyle{iclr2027_conference}

\newpage
\appendix
\section{Appendix}

\subsection{Construction Details}\label{app:construction}

\textbf{Passage benchmarks.} QuALITY articles carry 13 to 20 questions and
RACE passages 3 to 5. Within an article the questions are shuffled with a
fixed seed and cut into consecutive groups of $N$, $N{=}5$ for QuALITY and
$N{=}4$ for RACE. A remainder of fewer than $N$ questions is left out, and a
group in which two questions share a correct answer is skipped. On QuALITY
263 of the 265 articles contribute at least one group, and RACE yields 95
groups.

\textbf{No-context filter.} The three screening models are Gemini 3.6 Flash,
Gemini 3.1 Flash-Lite, and Gemma 4 26B. Each answers the original
multiple-choice items with the passage deleted, and a question is dropped
when two of the three answer it correctly. Of 3{,}972 QuALITY questions this
removes 49.7\%, against 15.6\% expected from guessing, and 28.8\% are
answered by all three. On the retained questions, Qwen3-8B, which the filter
never saw, scores 0.33 without the passage against a floor of 0.25. The
filter keeps the questions its screening models miss without the passage,
so their own no-context scores on the retained questions are low by
construction, and Table~\ref{tab:context} reports QuALITY over the other
eight models. The same screen marks 64.4\% of RACE questions, which leaves
too few per passage to form groups, so RACE is kept unfiltered and its
passage dependence is reported in Table~\ref{tab:context}.

\textbf{Topic benchmarks.} MMLU-Pro groups are formed within its 14
categories, GPQA within its subdomains, and C-Eval within its subjects, by a
seeded shuffle within each category.

\textbf{MMMU-Pro.} We use the standard configuration with ten options,
1{,}730 items in 30 subjects. Items whose options are themselves images are
dropped, 66 of them, since an image option cannot be listed beside another
question's text answers. 542 groups and 1{,}626 questions remain. Questions
cite their images inline and number them per question, so pooling renumbers
the citations into one sequence and orders the images to match.

\textbf{MedXpertQA-MM.} The test split has 2{,}000 questions with five
options and one to five images each, grouped by body system. The question
text embeds its options, which are removed from it, and a citation of each
image is prepended. Items whose options name a figure are dropped for the
same reason as MMMU-Pro's image options, 82 of them, leaving 1{,}918
questions in 635 groups.

\textbf{Video-MME.} Each video is resized to 448 pixels on the long side, and given to the model
as an ordered image sequence. The three questions of each video form one
group.

\textbf{Labels.} Options are labelled A to Z and then AA, AB, and so on, so
pools above 26 options are legal. On MMMU-Pro two-letter labels cover 7.7\%
of the answer slots, and on both benchmarks with $M{=}30$ their effect on
pooled accuracy is at most 0.02 and changes no ranking.

\subsection{Protocol}\label{app:protocol}

\textbf{Decoding.} All models decode greedily. Thinking is disabled through
the chat template where one exists. The gpt-oss models cannot disable
reasoning, so they run at the lowest reasoning effort and only their final
answer is parsed. Local models are served with vLLM 0.11, and a later release
for Qwen3.5, and decode under an output grammar that admits one line per
question in the requested form.

\textbf{Scoring.} A letter assigned to two questions is scored wrong for
both, since at most one of them can be right, and an unparsable slot is
scored wrong. With the full pool, 2.3 to 3.0\% of groups contain a repeated
letter.

\subsection{Per-model Results}\label{app:permodel}

\begin{table}[h]
\centering
\small
\caption{QuALITY, $N{=}5$, $M{=}20$, 300 groups.
Unrelated: the pool with each wrong option replaced by a correct answer of
another group (Appendix~\ref{app:controls}).}
\label{tab:quality}
\setlength{\tabcolsep}{4pt}
\begin{tabular}{lccccc}
\toprule
\textbf{Model} & MCQ & pooled & MCQ all-correct & pooled all-correct & unrelated \\
\midrule
Gemini 3.8 Flash      & 0.955 & 0.923 & 0.787 & \textbf{0.683} & 0.985 \\
Gemini 3.6 Flash      & 0.916 & 0.873 & 0.643 & 0.520 & 0.975 \\
Gemini 3.1 Flash-Lite & 0.877 & 0.787 & 0.513 & 0.317 & 0.937 \\
Gemma 4 26B           & 0.863 & 0.746 & 0.490 & 0.233 & 0.913 \\
Qwen3-32B             & 0.827 & 0.690 & 0.380 & 0.173 & 0.903 \\
Gemma 3 27B           & 0.808 & 0.669 & 0.367 & 0.160 & 0.897 \\
gpt-oss-120b          & 0.766 & 0.562 & 0.283 & 0.073 & 0.859 \\
Gemma 3 12B           & 0.731 & 0.583 & 0.217 & 0.073 & 0.854 \\
Qwen3-8B              & 0.719 & 0.571 & 0.227 & 0.080 & 0.797 \\
gpt-oss-20b           & 0.712 & 0.474 & 0.207 & 0.033 & 0.808 \\
Qwen2.5-7B            & 0.704 & 0.467 & 0.200 & 0.037 & 0.761 \\
\bottomrule
\end{tabular}
\end{table}

\begin{table}[h]
\centering
\small
\caption{MMLU-Pro, $N{=}3$, $M{=}30$, all nine wrong options pooled, groups by
category, 3{,}956 groups.}
\label{tab:mmlupro}
\begin{tabular}{lcccc}
\toprule
\textbf{Model} & MCQ & pooled & MCQ all-correct & pooled all-correct \\
\midrule
Gemini 3.8 Flash      & 0.898 & 0.886 & 0.743 & \textbf{0.714} \\
Gemini 3.6 Flash      & 0.776 & 0.722 & 0.501 & 0.427 \\
gpt-oss-120b          & 0.765 & 0.687 & 0.498 & 0.388 \\
Gemma 4 26B           & 0.725 & 0.518 & 0.406 & 0.185 \\
Gemini 3.1 Flash-Lite & 0.691 & 0.597 & 0.377 & 0.276 \\
gpt-oss-20b           & 0.681 & 0.552 & 0.368 & 0.236 \\
Qwen3-32B             & 0.561 & 0.442 & 0.225 & 0.132 \\
Gemma 3 27B           & 0.478 & 0.350 & 0.160 & 0.082 \\
Qwen3-8B              & 0.427 & 0.314 & 0.123 & 0.058 \\
Qwen2.5-7B            & 0.426 & 0.250 & 0.110 & 0.032 \\
Gemma 3 12B           & 0.424 & 0.312 & 0.122 & 0.060 \\
\bottomrule
\end{tabular}
\end{table}

\begin{table}[h]
\centering
\small
\caption{RACE, $N{=}4$, $M{=}16$, 95 groups. No passage: multiple-choice
accuracy with the passage removed.}
\label{tab:race}
\begin{tabular}{lccccc}
\toprule
\textbf{Model} & MCQ & pooled & MCQ all-correct & pooled all-correct & no passage \\
\midrule
Gemini 3.8 Flash      & 0.963 & 0.929 & 0.863 & \textbf{0.747} & 0.674 \\
Gemini 3.6 Flash      & 0.958 & 0.863 & 0.842 & 0.589 & 0.561 \\
Gemini 3.1 Flash-Lite & 0.937 & 0.821 & 0.768 & 0.484 & 0.484 \\
Gemma 4 26B           & 0.918 & 0.768 & 0.726 & 0.411 & 0.482 \\
Qwen3-32B             & 0.916 & 0.795 & 0.716 & 0.484 & 0.542 \\
Gemma 3 27B           & 0.897 & 0.745 & 0.621 & 0.389 & 0.484 \\
Qwen2.5-7B            & 0.895 & 0.637 & 0.642 & 0.221 & 0.534 \\
gpt-oss-120b          & 0.882 & 0.755 & 0.632 & 0.400 & 0.500 \\
Gemma 3 12B           & 0.882 & 0.716 & 0.621 & 0.337 & 0.497 \\
gpt-oss-20b           & 0.879 & 0.676 & 0.632 & 0.326 & 0.497 \\
Qwen3-8B              & 0.866 & 0.689 & 0.589 & 0.305 & 0.471 \\
\bottomrule
\end{tabular}
\end{table}

\begin{table}[h]
\centering
\small
\caption{GPQA, $N{=}5$, $M{=}20$, groups by subdomain, 83 groups.}
\label{tab:gpqa}
\begin{tabular}{lcccc}
\toprule
\textbf{Model} & MCQ & pooled & MCQ all-correct & pooled all-correct \\
\midrule
Gemini 3.8 Flash      & 0.887 & 0.812 & 0.554 & \textbf{0.349} \\
gpt-oss-120b          & 0.655 & 0.508 & 0.193 & 0.072 \\
Gemini 3.6 Flash      & 0.648 & 0.566 & 0.133 & 0.072 \\
gpt-oss-20b           & 0.557 & 0.407 & 0.072 & 0.060 \\
Gemini 3.1 Flash-Lite & 0.542 & 0.453 & 0.036 & 0.000 \\
Gemma 4 26B           & 0.511 & 0.376 & 0.036 & 0.024 \\
Qwen3-32B             & 0.424 & 0.354 & 0.012 & 0.012 \\
Qwen3-8B              & 0.388 & 0.272 & 0.012 & 0.000 \\
Gemma 3 27B           & 0.357 & 0.308 & 0.000 & 0.000 \\
Qwen2.5-7B            & 0.316 & 0.234 & 0.000 & 0.000 \\
Gemma 3 12B           & 0.311 & 0.299 & 0.000 & 0.000 \\
\bottomrule
\end{tabular}
\end{table}

\begin{table}[h]
\centering
\small
\caption{C-Eval, in Chinese, $N{=}5$, $M{=}20$, groups by subject, 237 groups.
Rank correlation 0.88.}
\label{tab:ceval}
\begin{tabular}{lcccc}
\toprule
\textbf{Model} & MCQ & pooled & MCQ all-correct & pooled all-correct \\
\midrule
Gemini 3.8 Flash      & 0.944 & 0.916 & 0.755 & \textbf{0.671} \\
Gemini 3.6 Flash      & 0.890 & 0.840 & 0.595 & 0.464 \\
Gemini 3.1 Flash-Lite & 0.847 & 0.755 & 0.464 & 0.295 \\
Qwen3-32B             & 0.840 & 0.700 & 0.451 & 0.241 \\
Qwen2.5-7B            & 0.762 & 0.494 & 0.300 & 0.021 \\
Qwen3-8B              & 0.752 & 0.552 & 0.312 & 0.084 \\
Gemma 4 26B           & 0.734 & 0.606 & 0.241 & 0.089 \\
gpt-oss-120b          & 0.732 & 0.613 & 0.291 & 0.135 \\
gpt-oss-20b           & 0.684 & 0.493 & 0.219 & 0.055 \\
Gemma 3 27B           & 0.619 & 0.494 & 0.135 & 0.046 \\
Gemma 3 12B           & 0.580 & 0.440 & 0.089 & 0.017 \\
\bottomrule
\end{tabular}
\end{table}

\begin{table}[h]
\centering
\small
\caption{MMMU-Pro, $N{=}3$, $M{=}30$, 542 groups. Unrelated: as in
Table~\ref{tab:quality}. No image: multiple-choice accuracy with the images
removed, against a floor of 0.10.}
\label{tab:mmmupro}
\setlength{\tabcolsep}{4pt}
\begin{tabular}{lcccccc}
\toprule
\textbf{Model} & MCQ & pooled & MCQ all-correct & pooled all-correct & unrelated & no image \\
\midrule
Qwen3.5-27B    & 0.513 & \textbf{0.411} & 0.162 & \textbf{0.090} & -- & 0.330 \\
Qwen3-VL-32B   & 0.472 & 0.363 & 0.140 & 0.061 & 0.870 & 0.299 \\
InternVL3-14B  & 0.441 & 0.308 & 0.107 & 0.042 & 0.823 & 0.285 \\
InternVL3-8B   & 0.396 & 0.221 & 0.092 & 0.017 & 0.734 & 0.262 \\
Qwen3-VL-8B    & 0.394 & 0.288 & 0.074 & 0.035 & 0.811 & 0.267 \\
Qwen2.5-VL-7B  & 0.378 & 0.239 & 0.089 & 0.017 & 0.775 & 0.231 \\
Gemma 3 27B    & 0.361 & 0.271 & 0.076 & 0.028 & 0.820 & 0.227 \\
Qwen3.5-4B     & 0.333 & 0.269 & 0.066 & 0.044 & -- & 0.238 \\
Gemma 3 12B    & 0.325 & 0.248 & 0.070 & 0.035 & 0.751 & 0.216 \\
\bottomrule
\end{tabular}
\end{table}

\begin{table}[h]
\centering
\small
\caption{Video-MME, $N{=}3$, $M{=}12$, one group per video, 897 groups. No
video: multiple-choice accuracy without the frames.}
\label{tab:videomme}
\setlength{\tabcolsep}{4pt}
\begin{tabular}{lccccc}
\toprule
\textbf{Model} & MCQ & pooled & MCQ all-correct & pooled all-correct & no video \\
\midrule
Qwen3.5-27B    & 0.690 & \textbf{0.628} & 0.356 & \textbf{0.281} & 0.449 \\
InternVL3-14B  & 0.650 & 0.563 & 0.294 & 0.226 & 0.426 \\
Qwen3-VL-32B   & 0.647 & 0.595 & 0.291 & 0.251 & 0.430 \\
Gemma 3 27B    & 0.627 & 0.548 & 0.263 & 0.201 & 0.435 \\
InternVL3-8B   & 0.620 & 0.483 & 0.262 & 0.144 & 0.384 \\
Qwen3-VL-8B    & 0.609 & 0.537 & 0.240 & 0.195 & 0.418 \\
Gemma 3 12B    & 0.592 & 0.499 & 0.226 & 0.156 & 0.392 \\
Qwen2.5-VL-7B  & 0.577 & 0.473 & 0.200 & 0.132 & 0.373 \\
Qwen3.5-4B     & 0.561 & 0.504 & 0.186 & 0.168 & 0.377 \\
\bottomrule
\end{tabular}
\end{table}

\begin{table}[h]
\centering
\small
\caption{MedXpertQA-MM, $N{=}3$, $M{=}15$, groups by body system, 635
groups. False answer: pooled, with one of the three answers removed per
group.}
\label{tab:medxpertqa}
\setlength{\tabcolsep}{4pt}
\begin{tabular}{lccccc}
\toprule
\textbf{Model} & MCQ & pooled & MCQ all-correct & pooled all-correct & false answer \\
\midrule
Qwen3.5-27B    & 0.429 & \textbf{0.349} & 0.063 & \textbf{0.044} & 0.997 \\
Qwen3-VL-32B   & 0.298 & 0.259 & 0.028 & 0.017 & 0.984 \\
Qwen3.5-4B     & 0.282 & 0.225 & 0.030 & 0.013 & 0.995 \\
InternVL3-14B  & 0.255 & 0.191 & 0.016 & 0.006 & 0.983 \\
Qwen3-VL-8B    & 0.252 & 0.205 & 0.024 & 0.013 & 0.976 \\
InternVL3-8B   & 0.241 & 0.184 & 0.013 & 0.008 & 0.981 \\
Gemma 3 27B    & 0.240 & 0.215 & 0.013 & 0.011 & 0.970 \\
Gemma 3 12B    & 0.234 & 0.203 & 0.011 & 0.006 & 0.965 \\
Qwen2.5-VL-7B  & 0.227 & 0.188 & 0.016 & 0.003 & 0.964 \\
\bottomrule
\end{tabular}
\end{table}

\subsection{Controls}\label{app:controls}

\textbf{Unrelated pool.} Each group keeps its questions, correct answers,
prompt, and pool size, and its wrong options are replaced by correct answers
of other groups. Every text model scores at or above its multiple-choice
accuracy on QuALITY (Table~\ref{tab:quality}), and on MMMU-Pro seven vision
models score 0.73 to 0.87 against 0.33 to 0.47 under multiple choice
(Table~\ref{tab:mmmupro}). Reading many questions and many options and
returning an assignment therefore costs no accuracy.

\textbf{Cross-domain groups.} The questions of a same-context build are
regrouped so that no two questions in a group share a passage (RACE and
QuALITY) or a subdomain (GPQA), and a group mixes numeric and textual option
sets where it can. Each question keeps its own options, and on passage
benchmarks its passage, labelled in the prompt (Appendix~\ref{app:prompts}).
A question that does not complete a group is left out. The regrouped builds
hold 95 groups on RACE, every question of its pooled build, 67 on GPQA, and
470 on QuALITY. Table~\ref{tab:xd} gives each model's accuracy on them next to
its multiple-choice and pooled accuracy from Table~\ref{tab:text}.

\begin{table}[h]
\centering
\small
\caption{Cross-domain groups. Accuracy under multiple choice / pooled /
cross-domain.}
\label{tab:xd}
\setlength{\tabcolsep}{4pt}
\begin{tabular}{lccc}
\toprule
\textbf{Model} & RACE & GPQA & QuALITY \\
\midrule
Qwen3-32B    & 0.916 / 0.795 / 0.842 & 0.424 / 0.354 / 0.358 & 0.827 / 0.690 / 0.457 \\
Gemma 3 27B  & 0.897 / 0.745 / 0.805 & 0.357 / 0.308 / 0.328 & 0.808 / 0.669 / 0.428 \\
Gemma 3 12B  & 0.882 / 0.716 / 0.782 & 0.311 / 0.299 / 0.325 & 0.731 / 0.583 / 0.370 \\
Qwen3-8B     & 0.866 / 0.689 / 0.750 & 0.388 / 0.272 / 0.313 & 0.719 / 0.571 / 0.368 \\
Qwen2.5-7B   & 0.895 / 0.637 / 0.695 & 0.316 / 0.234 / 0.263 & 0.704 / 0.467 / 0.314 \\
\bottomrule
\end{tabular}
\end{table}

\textbf{Own-set fraction.} On the QuALITY pool, 82 to 97\% of a model's picks
fall inside the question's own four options, against 20\% by chance. For
Gemini 3.6 Flash, accuracy within its own options is 0.902 against 0.916
under multiple choice, so its drop comes from picking another question's
answer.

\textbf{Group size.} The five local models on one QuALITY build of 414
groups, each $N$ a fresh chunking of the same articles. All-correct rate and
accuracy at $N{=}3,4,5,6$: Qwen3-32B 0.348, 0.246, 0.167, 0.090 and 0.695,
0.678, 0.657, 0.639. Gemma 3 27B 0.316, 0.208, 0.104, 0.059 and 0.660,
0.645, 0.627, 0.592. Gemma 3 12B 0.208, 0.130, 0.068, 0.034 and 0.576, 0.569,
0.548, 0.522. Qwen3-8B 0.206, 0.138, 0.058, 0.025 and 0.553, 0.542, 0.516,
0.495. Qwen2.5-7B 0.103, 0.065, 0.012, 0.000 and 0.458, 0.462, 0.442, 0.415.

\subsection{Abstention Details}\label{app:abstention}

\textbf{Multiple choice with an instruction.} On the items of
Table~\ref{tab:abstention}, deleting the answer from the item and permitting
none in the instruction gives false-answer rates of 0.120, 0.173, 0.300,
0.497, 0.575, 0.807, 0.777, 0.975, 0.930, 0.572, and 0.577 for the eleven
models in table order, between the pooled and the none-option rates for
every model.

\textbf{All-correct with abstentions.} Requiring every present answer to be
assigned and every removed one to be declined gives 0.300 for Gemini 3.8
Flash, 0.237 for Gemini 3.6 Flash, 0.067 for Gemma 4, 0.063 for Flash-Lite,
0.003 for Gemma 3 27B and gpt-oss-120b, and 0 for the remaining five models.

\textbf{Other benchmarks.} Table~\ref{tab:race_abst} repeats the three
formats on RACE, where two of four answers are removed per group, and
Table~\ref{tab:other_abst} gives the pooled rates on the topic benchmarks.
Gemini 3.8 Flash has the lowest false-answer rate on every benchmark, 0.17
on QuALITY, 0.21 on RACE, 0.24 on C-Eval, 0.31 on MMLU-Pro, 0.38 on GPQA,
and 0.54 on MedXpertQA-MM, and the Qwen models answer 90 to 100\% of the
unanswerable questions on every benchmark.

\begin{table}[h]
\centering
\small
\caption{Abstention on RACE, $p{=}0.5$, 95 groups, identical questions and
removed answers in every column.}
\label{tab:race_abst}
\begin{tabular}{lccccc}
\toprule
 & \multicolumn{3}{c}{False answer} & \multicolumn{2}{c}{False abstention} \\
\textbf{Model} & Pooled & MCQ+instruction & MCQ+none & Pooled & MCQ+none \\
\midrule
Gemini 3.8 Flash      & \textbf{0.211} & \textbf{0.100} & \textbf{0.095} & 0.032 & 0.105 \\
Gemini 3.6 Flash      & 0.221 & 0.121 & 0.111 & 0.105 & 0.084 \\
Gemini 3.1 Flash-Lite & 0.532 & 0.184 & 0.179 & 0.032 & 0.074 \\
Gemma 4 26B           & 0.511 & 0.279 & 0.268 & 0.116 & 0.089 \\
Gemma 3 27B           & 0.737 & 0.574 & 0.463 & 0.037 & 0.042 \\
Gemma 3 12B           & 0.926 & 0.721 & 0.432 & 0.032 & 0.121 \\
Qwen3-32B             & 0.968 & 0.495 & 0.500 & 0.005 & 0.037 \\
Qwen3-8B              & 0.984 & 0.916 & 0.421 & 0.005 & 0.063 \\
Qwen2.5-7B            & 0.932 & 0.816 & 0.189 & 0.021 & 0.211 \\
gpt-oss-120b          & 0.521 & 0.284 & 0.237 & 0.042 & 0.116 \\
gpt-oss-20b           & 0.500 & 0.311 & 0.268 & 0.063 & 0.079 \\
\bottomrule
\end{tabular}
\end{table}

\begin{table}[h]
\centering
\small
\caption{Pooled false-answer rate on the topic benchmarks, $p{=}0.4$, with
the false-abstention rate in parentheses.}
\label{tab:other_abst}
\begin{tabular}{lccc}
\toprule
\textbf{Model} & MMLU-Pro & GPQA & C-Eval \\
\midrule
Gemini 3.8 Flash      & \textbf{0.308} (0.057) & \textbf{0.378} (0.069) & \textbf{0.243} (0.030) \\
Gemini 3.6 Flash      & 0.573 (0.052) & 0.677 (0.106) & 0.377 (0.037) \\
Gemini 3.1 Flash-Lite & 0.842 (0.018) & 0.951 (0.016) & 0.813 (0.010) \\
Gemma 4 26B           & 0.823 (0.058) & 0.921 (0.037) & 0.864 (0.048) \\
gpt-oss-120b          & 0.617 (0.047) & 0.720 (0.159) & 0.743 (0.045) \\
gpt-oss-20b           & 0.617 (0.120) & 0.604 (0.350) & 0.732 (0.133) \\
Gemma 3 27B           & 0.915 (0.035) & 0.970 (0.024) & 0.915 (0.040) \\
Gemma 3 12B           & 0.930 (0.043) & 0.963 (0.037) & 0.968 (0.014) \\
Qwen3-32B             & 0.966 (0.009) & 0.994 (0.004) & 0.962 (0.003) \\
Qwen3-8B              & 0.991 (0.007) & 0.994 (0.020) & 0.998 (0.000) \\
Qwen2.5-7B            & 0.917 (0.062) & 0.927 (0.110) & 0.904 (0.020) \\
\bottomrule
\end{tabular}
\end{table}

\subsection{Ambiguity Check}\label{app:validity}

The screening models are Gemini 3.6 Flash, Gemini 3.1 Flash-Lite, and Gemma
4 26B, with the prompt of Appendix~\ref{app:prompts}. A question is flagged
when two of the three name another option, a reply that cannot be parsed
counts as no flag, and a group is kept when none of its questions is
flagged. On QuALITY the check flags 113 of 1{,}500 questions (7.5\%), 36
naming another question's correct answer and 77 a wrong option, touching 83
of 300 groups, so 72.3\% of groups are kept. On the kept groups Gemini 3.6
Flash's all-correct rate is 0.562 against 0.520 on all groups, and the gap
in false-answer rate between Gemini 3.6 Flash and Flash-Lite is 0.44.

\subsection{Prompts}\label{app:prompts}

Every setting fills one template with the item's text, and braces mark the
filled fields. Long lines are wrapped here for display only.

\textbf{Pooled, passage benchmarks.} The unrelated pool uses the same
template.
{\footnotesize
\begin{verbatim}
Read the passage, then match each question to its correct answer from the
candidate list.

Each candidate may be used at most once.

PASSAGE:
{passage}

QUESTIONS:
1. {question}
...

CANDIDATE ANSWERS:
A. {candidate}
...

Respond with one line per question in the form `<question number>: <letter>`,
and nothing else.
\end{verbatim}}

\textbf{Removed answers.} One line follows the rule line, and the last line
admits none. The second wording replaces the
added line with the one at the bottom.
{\footnotesize
\begin{verbatim}
Each candidate may be used at most once.
If a question's answer is NOT among the candidates, write "none" for it.
...
Respond with one line per question in the form `<question number>: <letter>`
or `<question number>: none`, and nothing else.

Some questions may have no correct answer in the list. Answer "none" for those.
\end{verbatim}}

\textbf{Topic, image, and video benchmarks.} There is no passage block, and
the template opens with the first line below, or with the second for video.
An image question cites its images as \texttt{<image k>}, and the images are
attached in that order.
{\footnotesize
\begin{verbatim}
Match each question to its correct answer from the candidate list.

The video is given as frames in order, attached before this text. Match
each question to its correct answer from the candidate list.
\end{verbatim}}

\textbf{Removed context.} Passage benchmarks drop the passage block and open
with the first lines below. Image benchmarks keep the text and attach no
image. Video opens with the last two lines.
{\footnotesize
\begin{verbatim}
Match each question to its correct answer from the candidate list.

Each candidate may be used at most once.
The source passage is not provided; answer as best you can.

The questions are about a video that is not provided; answer as best you can.
Match each question to its correct answer from the candidate list.
\end{verbatim}}

\textbf{Cross-domain groups.} GPQA uses the topic template. On passage
benchmarks every question brings its passage, and the question names it.
{\footnotesize
\begin{verbatim}
Read the passages, then match each question to its correct answer from the
candidate list.

Each candidate may be used at most once.

PASSAGES:
[Passage 1]
{passage of question 1}

[Passage 2]
{passage of question 2}
...

QUESTIONS:
1. [Passage 1] {question 1}
2. [Passage 2] {question 2}
...
\end{verbatim}}

\textbf{Multiple choice.} Every item keeps its own options in their original
order. Topic and image benchmarks open with the second line below and have
no passage block, and video opens with the third. Without the context,
passage benchmarks open with the fourth and video with the fifth, and image
benchmarks attach no image.
{\footnotesize
\begin{verbatim}
Read the passage and answer the question.

PASSAGE:
{passage}

QUESTION: {question}

OPTIONS:
A. {option}
...

Respond with only the letter of the correct option.

Answer the question.
The video is given as frames in order, attached before this text. Answer
the question about the video.
Answer the question. The source passage is not provided; answer as best you
can.
Answer the question about a video. The video is not provided; answer as best
you can.
\end{verbatim}}

\textbf{Multiple choice with a none option.} Three of the item's options, the
correct one among them when it is present, and the none option, shuffled.
{\footnotesize
\begin{verbatim}
Read the passage and answer the question.

PASSAGE:
{passage}

QUESTION: {question}

OPTIONS:
{three options and "None of these answers is correct.", shuffled}

One of the options states that none of the listed answers is correct. Choose
it if, and only if, that is the case.
Respond with only the letter of the correct option.
\end{verbatim}}

\textbf{Multiple choice with an instruction.} Three options drawn the same
way, without a printed none.
{\footnotesize
\begin{verbatim}
Read the passage and answer the question.

PASSAGE:
{passage}

QUESTION: {question}

OPTIONS:
{three options}

If the correct answer is NOT among the options, write "none".
Respond with only the letter of the correct option, or the word none.
\end{verbatim}}

\textbf{Ambiguity check.} The screener sees the pool of the question's group
with the correct answer revealed. On benchmarks without a passage, the
passage field reads ``(This benchmark has no passage. Judge from expert
knowledge of the subject.)''.
{\footnotesize
\begin{verbatim}
You are auditing a benchmark item for ambiguity.

PASSAGE:
{passage}

QUESTION: {question}

CANDIDATE ANSWERS:
A. {candidate}
...

The intended correct answer is {letter}.

Based only on the passage, is any OTHER candidate ALSO a fully correct answer
to this question? A candidate that is merely related or partially true does
not count. It must be a fully correct answer on its own.

Reply with exactly one token: the letter of such a candidate, or NO.
\end{verbatim}}

\end{document}